\documentclass{article}   

\PassOptionsToPackage{numbers,sort&compress}{natbib}

\usepackage[dblblindworkshop, final]{neurips_2025}
\usepackage{amsmath}
\usepackage{enumitem}
\usepackage{graphicx}
\usepackage{subcaption}
\usepackage{float}
\usepackage{hyperref}
\usepackage{siunitx}

\usepackage[utf8]{inputenc} 
\usepackage[T1]{fontenc}    
\usepackage{url}            
\usepackage{booktabs}       
\usepackage{amsfonts}       
\usepackage{nicefrac}       
\usepackage{microtype}      
\usepackage{xcolor}         

\title{SwiftSolve: A Self-Iterative, Complexity-Aware Multi-Agent Framework for Competitive Programming}

\author{
  Adhyayan Veer Singh\\
Algoverse Research
  \And
  Aaron Shen\\\
  Algoverse Research
  \And
  Brian Law\\
  Algoverse Research
  \And
  Ahmed Ismail\\
  UC Berkeley \\
  \And
  Jonas Rohweder\\
  TU Darmstadt\\
  Zuse School ELIZA\\
  \And
  Sean O'Brien\\
  Algoverse Research\\
  \And
  Kevin Zhu\\
  Algoverse Research\\
}

\workshoptitle{Efficient Reasoning}

\begin{document}

\maketitle


\begin{abstract}
Correctness alone is insufficient: LLM-generated programs frequently satisfy unit tests while violating contest time or memory budgets. We present \textit{SwiftSolve}, a complexity-aware multi-agent system for competitive programming that couples algorithmic planning with empirical profiling and complexity-guided repair. We frame competitive programming as a software environment where specialized agents act as programmers, each assuming roles such as planning, coding, profiling, and complexity analysis. A Planner proposes an algorithmic sketch; a deterministic Static Pruner filters high-risk plans; a Coder emits ISO~C++17; a Profiler compiles and executes candidates on a fixed input-size schedule to record wall time and peak memory; and a Complexity Analyst fits log--log growth (slope \(s\), \(R^2\)) with an LLM fallback to assign a complexity class and dispatch targeted patches to either the Planner or Coder. Agents communicate via typed, versioned JSON; a controller enforces iteration caps and diminishing returns, stopping. Evaluated on 26 problems (16 BigO(Bench), 10 Codeforces Div.~2), three seeds each (\(N=78\) runs) in a POSIX sandbox (2\,s / 256--512\,MB), SwiftSolve attains \textsc{pass@1} \(=\) 61.54\% (16/26) on the first attempt and \textsc{Solved@}\(\le\)3 \(=\) 80.77\% with marginal latency change (mean 11.96\,s \(\to\) 12.66\,s per attempt). Aggregate run-level success is 73.08\% at 12.40\,s mean. Failures are predominantly resource-bound, indicating inefficiency rather than logic errors as the principal barrier. Against a Claude Opus 4 single-agent baseline, SwiftSolve improves run-level success (73.1\% vs.\ 52.6\%) at \(\sim\)2\(\times\) runtime overhead (12.4\,s vs.\ 6.8\,s). Beyond correctness (\textsc{pass@k}), we report efficiency metrics (\textsc{eff@k} for runtime/memory, incidence of TLE / MLE, and complexity fit accuracy on BigO (Bench), demonstrating that profiling and complexity-guided replanning reduce inefficiency while preserving accuracy. Future studies would integrate comparisons against other multi-agent frameworks and include ablation studies. Our implementation is available at \url{https://github.com/jonasrohw/swiftsolve/.}
\end{abstract}

\section{Introduction}

Code generation is primarily used to automate software development by generating functional and context-aware code from natural language inputs \cite{hossain_llm-pros_2025}. Automated code generation can lead to enhanced productivity and a reduction in manual coding efforts \cite{sepidband_enhancing_2025}, resulting in more efficient systems. In recent years, multi-agent systems have emerged as a feasible alternative to single-agent frameworks in the domain of LLM-based code generation \cite{huang_agentcoder_2024, ishibashi_self-organized_2024}.

Recently, researchers have proposed multi-agent frameworks that allow LLMs to analyze, generate, and iteratively patch code \cite{huang_effilearner_2025, liu_lessons_2025, pan_codecor_2025, shinn_reflexion_2023, huang_agentcoder_2024}. CodeCoR \cite{pan_codecor_2025} and CodeSIM \cite{islam_codesim_2025} are such examples. CodeCoR consists of four LLM-based agents for prompting, task understanding, test case generation, as well as code repairing. Meanwhile, CodeSIM involves three LLM-based agents for planning, code generation, and debugging, with the inclusion of a verification step for the planning agent, mimicking how humans understand, visualize, and refine algorithms.

However, significant gaps remain. While many studies focus on optimizing the correctness of code \cite{ishibashi_self-organized_2024, zhang_towards_2024, huang_agentcoder_2024}, few studies focus on improving runtime or memory complexity, resulting in inefficient code that could lead to time limit exceeded (TLE) or memory limited exceeded (MLE) errors. Additionally, previous multi-agent frameworks iteratively test and patch code \cite{huang_effilearner_2025, pan_codecor_2025}, but never consider asymptotic performance, causing slow code generation. 

To address these gaps, we introduce SwiftSolve, a novel self-iterative multi-agent framework that can generate correct, complete, and efficient code. Specifically, SwiftSolve consists of 4 LLM-based agents: \emph{1) Planner Agent}, which creates an algorithmic sketch based off of a natural language prompt; \emph{2) Coder Agent}, which generates ISO C++17 code from the algorithmic sketch;\emph{3) Profiler Agent}, which profiles candidates on a fixed, deterministic input-size schedule; and \emph{4) Complexity Analyst Agent}, which will infer time and memory complexity, compares the estimate to contest constraints, and dispatches either a minor fix to the coder agent or an algorithmic overhaul to the planner agent.

In addition, we introduce the use of a lightweight static pruner \cite{pan_codecor_2025} designed to filter out inefficient \emph{plans} from the Planner before code generation and any expensive LLM calls. This pruner communicates through typed JSON messages and determines whether to return to the Planner (for large errors) or the Coder to allow for cheap, efficient iterations.

We evaluate our framework on two datasets—BigO(Bench)~\cite{chambon_bigobench_2025} and Codeforces Div.~2~\cite{codeforces-2025}. On a snapshot covering $N{=}78$ task–seed trials (26 tasks × 3 seeds), {SwiftSolve} attains \textsc{pass@1} defined on the first attempt (\texttt{replan\_0}) of $61.54\%$ ($16/26$). We report \textsc{pass@k} and efficiency (\textsc{eff@k}) metrics from our sandboxed profiler and include a GPT\!-\!4.1 Single-Agent baseline.

In summary, our contributions are two-fold:
\begin{itemize}
\item We propose SwiftSolve, a novel self-iterative multi-agent framework consisting of 4 distinct agents (i.e, planner agent, coder agent, profiler agent, and complexity analyst) for complete, correct, and effective code generation. A lightweight static pruner filters out inefficient \emph{plans} from the Planner before code generation, routing back to the Planner for large errors or allowing the Coder to proceed.
\item We demonstrate the effectiveness of SwiftSolve against GPT-4.1 and Claude 4 Opus through extensive comparison. Experimental results show that SwiftSolve significantly outperforms the baseline.
\end{itemize}

\section{Related work}
\paragraph{Self-Reflective and Multi-Agent Code Generation}
There have been multiple advancements in self-iterative multi-agent frameworks for code generation and correctness \cite{sepidband_enhancing_2025, nichols_performance-aligned_2024, qiu_how_2025, wei_evaluating_2025, hong_metagpt_2024, hossain_llm-pros_2025}. For example, Effilearner \cite{huang_effilearner_2025}, a self-optimizing framework for code generation, consists of 3 main parts: code generation, overhead profiling, and code refinement. This framework also incorporates a feedback loop designed to decrease code complexity and improve performance. Extensive experiments have corroborated Effilearner's effectiveness on Effibench, HumanEval, and MBPP, reducing overhead and increasing the efficiency of code generation. Unlike Effilearner, SwiftSolve employs four distinct agents, each assigned a specialized role to ensure expertise and efficiency in its designated task.

Similarly, subsequent works such as Liu et. al \cite{liu_lessons_2025} used a team of LLM agents to learn from each other's successes and failures to improve code optimization tasks. The knowledge gained from each iteration is then demanded and deposited in a bank to be accessed by other agents, fostering a collaborative learning environment \cite{wei_evaluating_2025}. This lays the groundwork for future collaborative LLMs that don't consider an agent's complementary strengths a priori, but still contribute to the overall knowledge and effectiveness of the framework. Runtime speedups and resource efficiency metrics are then calculated to demonstrate code optimization. SwiftSolve, though mirroring the collaborative environment, additionally offers insight into runtime and memory complexity, ensuring framework robustness.
\paragraph{Role Specific LLMs}
The domain of role-based multi-agent frameworks, in which each agent has a specialized task, has become more prominent in recent years. For example, MapCoder \cite{islam2024mapcodermultiagentcodegeneration}, mimicking the human iteration cycle utilizes four agents for retrieval, planning, coding, and debugging. This state-of-the-art method has led to groundbreaking accuracy in code correctness and generation. Similarly, frameworks such as MetaGPT \cite{hong_metagpt_2024} and AgentCoder \cite{huang_agentcoder_2024} provide a name, profile, goal, and constraints for each agent to ensure speciality and specific context for their roles. In addition to their specializations, MetaGPT has each individual agent monitor the environment, collecting important observations such as messages from other agents. This ensures the collaborative nature of the framework, allowing agents to assist each other when needed.

\section{Methodology}
\label{sec:method}

\begin{figure}[H]
  \centering
\includegraphics[width=\linewidth,height=0.35\textheight,keepaspectratio]{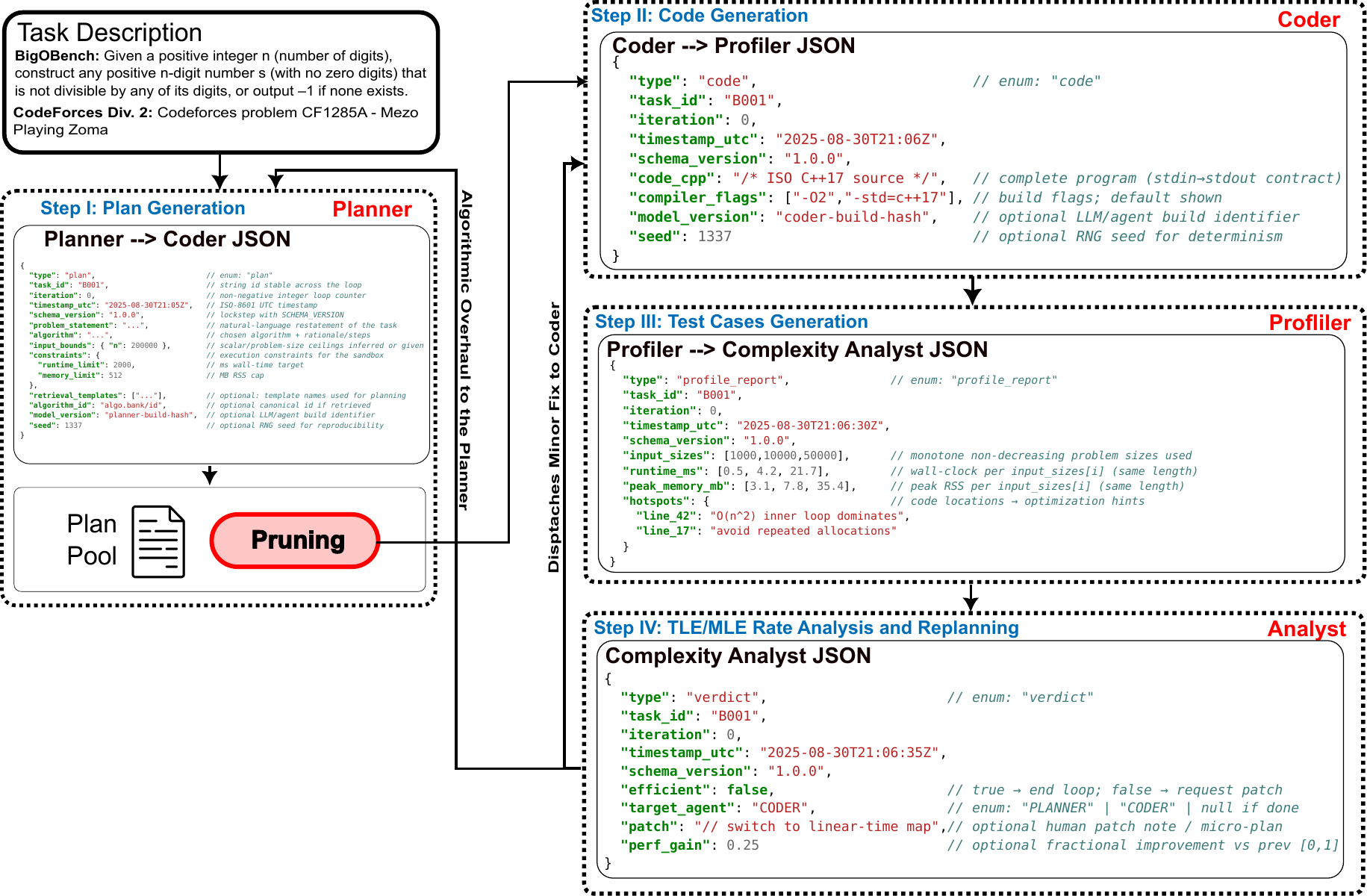}
  \caption{\textbf{SwiftSolve pipeline.} Natural-language prompt $\rightarrow$ Planner
  $\rightarrow$ Static Pruner $\rightarrow$ Coder $\rightarrow$ Profiler
  $\rightarrow$ Complexity Analyst with JSON feedback to the Coder (and optional
  Planner) until an efficient solution is reached.}
 \label{fig:pipeline}
\end{figure}

Our system is a modular, message-passing pipeline that co-optimizes code \emph{correctness} and \emph{asymptotic efficiency}. Figure~\ref{fig:pipeline} (overview) shows five components orchestrated by a controller: \textbf{Planner} $\to$ \textbf{Static Pruner} $\to$ \textbf{Coder} $\to$ \textbf{Profiler} $\to$ \textbf{Complexity Analyst} (with iterative feedback). Agents communicate solely via typed JSON envelopes backed by Pydantic~v2 \cite{pydantic_validation}. All messages carry a frozen header with \texttt{type}, \texttt{task\_id}, \texttt{iteration}, \texttt{timestamp\_utc}, and \texttt{schema\_version}~\texttt{"1.0.0"}.

\subsection{System architecture}
\label{sec:arch}

\paragraph{Typed interfaces.}
Agents exchange only versioned, typed JSON envelopes validated with Pydantic~v2. Every envelope includes an immutable header—\texttt{type}, \texttt{task\_id}, \texttt{iteration}, \texttt{timestamp\_utc}, \texttt{schema\_version}—and a schema-specific payload. Headers provide idempotence and ordering through the (\texttt{task\_id}, \texttt{iteration}) key, and the controller rejects messages with unknown fields or missing required fields to guarantee forward compatibility and auditability. All messages are durably logged and are treated as read-only artifacts once emitted.

\textit{PlanMessage} is the Planner’s machine-consumable contract to the rest of the system. It carries the problem statement in natural language, an algorithm-family label that specifies the intended approach at the level of asymptotic behavior (e.g., two-pointers vs divide–and–conquer), declared input bounds and contest constraints (time in milliseconds and memory in megabytes), and optional retrieval hints for templated subroutines. It also records provenance (\texttt{model\_version}, \texttt{seed}) and an optional algorithm identifier for downstream traceability. The Static Pruner consumes \textit{PlanMessage} to deterministically reject obviously risky plans under the stated bounds before any code generation; plans that pass are the only inputs the Coder is permitted to realize.

\textit{CodeMessage} transports a complete ISO~C++17 translation unit along with the compiler flags required to reproduce the binary that will be profiled. The Coder emits exactly one translation unit per attempt; a thin sanitizer may normalize headers and formatting to ensure successful compilation while preserving semantics. Provenance fields mirror the planner’s to enable per-seed determinism studies. The Profiler is the sole consumer of \textit{CodeMessage}; no other agent mutates code payloads, and any subsequent edits must be expressed as a new \textit{CodeMessage} at a higher iteration index.

\textit{ProfileReport} is the Profiler’s measurement artifact. It returns the deterministic input schedule used for probing, the corresponding vectors of wall-clock runtime and peak resident memory for each input size, and a lightweight hotspot map with textual notes on potential bottlenecks. Failures are represented in-place: timeouts or nonzero exits are encoded as missing or $+\infty$ entries for the affected sizes while allowing the remainder of the schedule to complete. The report additionally includes the slope and $R^2$ of a least-squares fit in log–log space and the resulting complexity label when available; these quantities are consumed by the Complexity Analyst and the controller for routing decisions, patch synthesis, and termination checks. Full schema definitions are deferred to the appendix.

\paragraph{Execution target and toolchain.}
All candidate programs are single-translation-unit ISO~C++17. The Profiler compiles with
\texttt{g++ -O2 -std=c++17 -march=native -ffast-math} (two attempts) and executes the binary under
\texttt{/usr/bin/time -v} to obtain elapsed wall-clock time and maximum resident set size (RSS).
Per-run process timeouts enforce liveness (default: \textbf{2\,s}).

\paragraph{Controller.}
The controller invokes agents, logs all JSON payloads, and enforces safety (agent-failure budget, iteration cap, diminishing-returns stop). Outcomes are one of:
\texttt{success}, \texttt{static\_prune\_failed}, or \texttt{agent\_failure} (budget exhausted or insufficient gain).

\subsection{Agent Roles}
\label{sec:roles}

\subsubsection{Planner}
The Planner (Claude~4~Opus) converts the natural-language task into a minimal, machine-consumable plan:
\texttt{algorithm} (e.g., ``two\_pointers''), integer \texttt{input\_bounds} (e.g., \texttt{\{"n": 100000\}}),
and integer \texttt{constraints} (e.g., \texttt{\{"runtime\_limit": 2000, "memory\_limit": 512\}}).
On replanning, the system prompt \emph{requires a different algorithmic family} and to address performance feedback.

\subsubsection{Static Pruner}
Before any LLM generation is expended on code, a deterministic gate rejects plans that are provably risky given the declared bounds.
Let $n=\texttt{input\_bounds["n"]}$ (default 0 if absent) and $\alpha=\texttt{algorithm}$ lowercased.
We reject if any of the following triggers fire:
\begin{enumerate}
  \item \textbf{While-loop multiplicity at large $n$:} $\texttt{count}(\alpha,\text{``while''}) > 2$ and $n \ge 10^5$.
  \item \textbf{Unbounded recursion at large $n$:} the substring ``recursion'' occurs in $\alpha$ and $n \ge 10^4$.
  \item \textbf{Sort-inside-loop pattern at moderate $n$:} a regex approximating ``\texttt{for ...: ... sort(...)}'' matches $\alpha$ and $n \ge 10^3$.
\end{enumerate}
If any rule triggers, the controller returns \texttt{static\_prune\_failed}. Otherwise the plan advances.

\subsubsection{Coder}
The Coder (GPT-4.1) emits a complete, compilable C++17 program. Two modes are supported:
\emph{(i) initial synthesis} from the plan alone, and
\emph{(ii) patch application} where the system prompt mandates incorporation of the Analyst’s optimization.
A thin sanitizer adds missing standard headers and repairs common formatting issues, but otherwise preserves the model’s output.

\subsubsection{Profiler}
Given a \texttt{CodeMessage}, the Profiler compiles and executes the candidate against a fixed input schedule and returns a \texttt{ProfileReport}.
\begin{itemize}
  \item \textbf{Compilation.} \texttt{g++ -O2 -std=c++17 -march=native -ffast-math}; one automatic retry on failure.
  \item \textbf{Input schedule.} Deterministic sizes $\langle 0,\,1,\,10^3,\,5\!\cdot\!10^3,\,10^4,\,5\!\cdot\!10^4,\,10^5\rangle$.
        The default generator writes a single integer line ``\texttt{n\textbackslash n}'' (task-specific generators are pluggable; see \S\ref{sec:dynprof}).
  \item \textbf{Telemetry.} \texttt{/usr/bin/time -v} provides wall-clock and max RSS; outputs are parsed via regex.
        Each per-$n$ execution inherits a process-level timeout (2\,s). On timeout or nonzero exit, that point is recorded as $+\infty$ in both series and the loop continues.
  \item \textbf{Hotspots.} A lightweight map records crash notes or hints; gprof-based attribution is stubbed and does not affect routing.
\end{itemize}

\subsubsection{Complexity Analyst}
The Analyst integrates regression-based fitting with an LLM fallback to assign a time-complexity class and to synthesize targeted optimizations.
Let $D=\{(n_i,t_i)\}$ be valid points where $t_i>0$ and finite.
\begin{enumerate}
  \item \textbf{Slope fitting.} We fit a line in log--log space between input size and runtime, and compute the slope \(s\) together with the coefficient of determination \(R^2\). The choice of logarithm base does not affect \(s\) or \(R^2\).
  \item \textbf{Ambiguity test.} Mark the curve ambiguous if any hold:
        $R^2<0.7$; $s\in(0.4,0.6)\cup(1.3,1.7)\cup(2.3,2.7)$; significant non-monotone noise (at least one $>10\%$ increase and one $>10\%$ decrease); $s<-0.5$ or $s>10$; or the size range $<10\times$.
  \item \textbf{Classification.} If unambiguous, map by thresholds:
        $s<0.5\Rightarrow\mathcal{O}(1)$,
        $s<1.5\Rightarrow\mathcal{O}(n)$,
        $s<2.5\Rightarrow\mathcal{O}(n^2)$,
        else $\mathcal{O}(n^k)$ (coarse bucket).
        If ambiguous, query GPT-4.1 for a \emph{single} label from
        $\{\mathcal{O}(1),\mathcal{O}(\log n),\mathcal{O}(n),\mathcal{O}(n\log n),\mathcal{O}(n^2),\mathcal{O}(n^3),\mathcal{O}(2^n),\mathcal{O}(n!)\}$ and normalize the response.
  \item \textbf{Efficiency decision and patching.}
        We set \texttt{efficient}~$\!=\!$~\texttt{true} iff the final class $\in\{\mathcal{O}(1),\mathcal{O}(\log n),\mathcal{O}(n),\mathcal{O}(n\log n)\}$.
        Otherwise, the Analyst emits \texttt{target\_agent=\text{CODER}} and a natural-language \texttt{patch} (e.g., replace nested loops with hash lookups; adopt two pointers; reduce allocations).
        Memory series are inspected in patch logic (e.g., flagging excessive growth for otherwise linear code) but do not currently gate \texttt{efficient}.
\end{enumerate}

\subsection{Dynamic Profiling}
\label{sec:dynprof}
We profile each candidate across a fixed, log-spaced schedule augmented with corner cases ($n\in\{0,1\}$). This provides stable variance and comparable growth curves across tasks.
The first version employs a \emph{task-agnostic} input generator that only writes $n$.
While this stresses raw iteration/recursion costs and common STL usage, it under-exercises branch-dependent worst cases for some problems (e.g., value distributions for two-pointers or pivot patterns in quicksort).
The API is designed for drop-in \emph{task-specific adversarial generators} (dense/sparse graphs, nearly-sorted arrays, pathological duplicates); integrating these does not change any upstream schema.

\subsection{Complexity Fitting}
\label{sec:fit}
Complexity fitting follows the \emph{Complexity Analyst} procedure in §\ref{sec:roles}. From the Profiler we obtain \(D=\{(n_i,t_i)\}\) and estimate the slope \(s\) and coefficient of determination \(R^2\) via least-squares in log–log space; class assignment and the LLM fallback for ambiguous fits are exactly as specified in §3.2.5. We retain \(D\), \(s\), \(R^2\), and the assigned class; these fields are emitted in \texttt{ProfileReport} and consumed by the Analyst and the judge. Peak memory (\textit{peak\_memory\_mb}) is recorded concurrently.

\subsection{Replanning Policy}
\label{sec:policy}
After iteration $k$ produces profile $\{(n_i,t_i^{(k)})\}$, we compute a fractional improvement on the largest input size:
\begin{equation}
\label{eq:gain}
\mathrm{gain}^{(k)} \;=\; \frac{t_{\max}^{(k-1)} - t_{\max}^{(k)}}{t_{\max}^{(k-1)}} \;\;,
\qquad
t_{\max}^{(k)} \equiv t_{i^\star}^{(k)}\text{ with }n_{i^\star}=\max_i n_i .
\end{equation}
If $\mathrm{gain}^{(k)} < \delta$, we terminate; otherwise we apply the Analyst’s \texttt{patch} in the next Coder call.
Routing is governed by \texttt{target\_agent}; the current implementation primarily targets the Coder, while the controller also supports Planner-directed replans that generate a \emph{different} algorithm; the budget is capped at three attempts per task (\texttt{replan\_0}, \texttt{\_1}, \texttt{\_2}).

\subsection{Termination Criteria and Safety Guards}
\label{sec:safety}
The loop stops under any of the following conditions:
\begin{enumerate}
  \item \textbf{Success:} the Analyst returns \texttt{efficient=true}.
  \item \textbf{Diminishing returns:} $\mathrm{gain}^{(k)} < \delta$ (configurable).
  \item \textbf{Iteration cap:} at most \textbf{3} iterations (controller default).
  \item \textbf{Static rejection:} the Static Pruner rejects the plan.
  \item \textbf{Agent-failure budget:} two caught exceptions across Planner/Coder/Profiler/Analyst abort the run.
\end{enumerate}
Process-level safety includes isolated temp workspaces, strict per-run timeouts (2\,s), capture of stdout/stderr, and graceful point-wise failure handling (recorded as $+\infty$ without crashing the loop).

\section{Experimental setup}\label{sec:setup}

\paragraph{Tasks and datasets.}
We evaluate on two sources comprising \textbf{26 tasks} total: \textbf{16} from BigO(Bench)~\cite{chambon_bigobench_2025} (binary\_search, linear\_scan, two\_pointers, hash\_lookup, prefix\_sum, sliding\_window, merge\_sort, quick\_sort, heap\_operations, bfs\_graph, dfs\_graph, dijkstra, dynamic\_programming, backtracking, matrix\_multiplication, segment\_tree) and \textbf{10} Codeforces Div.\,2 problems~\cite{codeforces-2025} (e.g., \textit{cf\_1285a\_maximum\_array}, \textit{cf\_1285b\_just\_eat\_it}, \textit{cf\_1285c\_fadi\_and\_lcm}, \textit{cf\_1285d\_dr\_evil\_underscores}). Unless otherwise stated, all analyses operate on this same fixed task list. The snapshot comprises \(N{=}78\) \emph{task–seed trials} (26 tasks × 3 seeds). Within each trial, the controller permits up to three attempts on the \emph{same} problem: \texttt{replan\_0}, \texttt{replan\_1}, \texttt{replan\_2} (i.e., at most two replans beyond the initial).

\paragraph{Environment and sandbox.}
Candidates are compiled and executed in a POSIX sandbox with per-task limits taken from the harness metadata. Runtime and memory limits mirror the source benchmarks for comparability: BigO(Bench) tasks use a 2\,s wall-clock budget and 512\,MB memory as specified by the harness \cite{chambon_bigobench_2025}; Codeforces Div.2 problems use the platform’s standard 2\,s and 256\,MB limits \cite{codeforces-2025}. We adopt the per-task limits from these sources without modification. Each run logs wall time, peak RSS, exit status, and (when available) per-input profiling traces (\texttt{input\_sizes}, \texttt{runtime\_ms}, \texttt{peak\_memory\_mb}) together with boolean \texttt{TLE}/\texttt{MLE} flags over the input grid.

\paragraph{Agent protocol.}
For each task we run up to three attempts with lightweight replanning between attempts (\texttt{replan\_0}, \texttt{replan\_1}, \texttt{replan\_2}). Each attempt yields one compiled candidate judged on the hidden test harness. A task is marked \texttt{success} if any of its attempts pass all checks; otherwise \texttt{agent\_failure}.

\paragraph{Metrics.}
\textsc{pass@1} refers strictly to the first attempt (\texttt{replan\_0}) on each task.
\textbf{Solved@$\le k$} is the fraction of the same task list solved within at most $k$ attempts (cumulative over \texttt{replan\_0}, \texttt{\_1}, \texttt{\_2} on the identical problems).
\textsc{eff@k} (runtime/memory) is pass@k under an efficiency predicate: no TLE/MLE anywhere on the profiled grid and a fitted runtime curve that extrapolates under the task limits at $n_{\max}$. We primarily report \textsc{eff@1} for time and memory. When more than $k$ attempts exist, we use the unbiased order-statistics estimator from ENAMEL~\cite{qiu2025enamel}.

\textbf{TLE/MLE rate:} the fraction of probed input sizes that triggered time limit exceeded (TLE)/memory limit exceeded (MLE) for a candidate; summarized per task and in aggregate.

\textbf{Complexity-fit accuracy (BigO(Bench)).} For solutions with per-size profiles, we assign a label via the slope-threshold procedure with an LLM fallback exactly as specified in §3.2.5, and report the fraction matching the benchmark’s ground-truth Big-O.

\paragraph{Evaluation summary.}
Table~\ref{tab:overall} reports headline numbers. We define \textsc{pass@1} on the first attempt only (\texttt{replan\_0}); under this definition \textsc{pass@1} is $16/26{=}61.54\%$. Replanning indices (\texttt{replan\_0}, \texttt{\_1}, \texttt{\_2}) are successive attempts on the \emph{same} problems under a fixed seed. The sweep uses $26$ tasks × $3$ seeds = \textbf{78} runs in total; the mean per-run duration is $12.40 s$.

\begin{table}[h]
\centering
\small
\caption{Aggregate run-level statistics. Cumulative wall time: 16.12 minutes.}
\vspace{1.0em}
\begin{tabular}{lrrrr}
\toprule
& Total runs & Successes & \textsc{Solved@$\le$3} & Avg. time / run \\
\midrule
Overall & 78 & 57 & 73.08\% & 12.40 s \\
\bottomrule
\end{tabular}

\label{tab:overall}
\end{table}

\paragraph{Replanning}
Table~\ref{tab:replanning} summarizes outcomes by attempt index. Success rate improves monotonically with additional replanning budget, with a modest change in latency.

\begin{table}[ht]
    \centering
    \small
    \caption{Effect of replanning budget. ``Tasks'' counts the same 26 problems evaluated at each attempt index; percentages are cumulative solved rates over that fixed set.}
    \vspace{1.0em}
    \begin{tabular}{lrrr}
        \toprule
        Attempt & Tasks & \textsc{Solved@$\le k$} & Avg. time / attempt \\
        \midrule
        \texttt{replan\_0} & 26 & 61.54\% & 11.96 s \\
        \texttt{replan\_1} & 26 & 76.92\% & 12.58 s \\
        \texttt{replan\_2} & 26 & 80.77\% & 12.66 s \\
        \bottomrule
    \end{tabular}
    \label{tab:replanning}
\end{table}

\paragraph{Per-source breakdown.}
Table~\ref{tab:by-source} shows performance by dataset source. BigO(Bench) runs are faster on average and exhibit higher acceptance than the Codeforces-style subset.

\begin{table}[h]
\centering
\small
\caption{Performance by dataset source.}
\vspace{1.0em}
\begin{tabular}{lrrr}
\toprule
Source & Count & \textsc{Solved@$\le$3} & Avg. time / run \\
\midrule
BigO(Bench) & 49 & 77.55\% & 10.30 s \\
Codeforces subset & 29 & 65.52\% & 15.95 s \\
\bottomrule
\end{tabular}

\label{tab:by-source}
\end{table}

\paragraph{Computation of metrics from logs.}
All results in Tables~\ref{tab:overall}--\ref{tab:by-source} are computed directly from the evaluation JSON (\texttt{evaluation\_summary}, \texttt{replan\_analysis}, and \texttt{research\_insights}). For \textsc{eff@k} and complexity-fit, we use the per-run profiling arrays (\texttt{input\_sizes}, \texttt{runtime\_ms}, \texttt{peak\_memory\_mb}) and the sandbox limits recorded under \texttt{constraints}. TLE/MLE rates are averaged over the Boolean flag vectors provided for each attempt. We release the raw JSON alongside the paper for exact recomputation.


\section{Results}
\subsection{Main Performance Comparison}
\label{sec:mainperform}
In total 78 runs were executed (Table~\ref{tab:overall}). Of these, 57 runs completed successfully, yielding an overall success rate of 73.1\%. The mean execution time per run was 12.4 seconds. Table~\ref{tab:overall} summarizes these aggregate metrics. The two datasets show different results: BigO(Bench) problems (49 runs) achieved ~77.6\% success with a 10.3 s average time, whereas Codeforces problems (29 runs) had ~65.5\% success with a 15.95 s average (Table~\ref{tab:by-source}, Figure~\ref{fig:example1}). These figures indicate that performance varied by task type.

Dataset breakdown: BigO(Bench) – 49 runs, 77.6\% success, 10.3 s avg. Codeforces – 29 runs, 65.5\% success, 15.95 s avg.

All 21 failed runs were attributable to efficiency problems rather than coding errors. Several plans were rejected immediately by the static pruner (e.g., excessive nested loops), and the rest terminated after the maximum iterations when the profiler continued to report TLE/MLEs.

\subsection{Runtime Curves}
\begin{figure}[h]
    \centering
    \begin{subfigure}{0.48\textwidth}
        \centering
        \includegraphics[width=\linewidth]{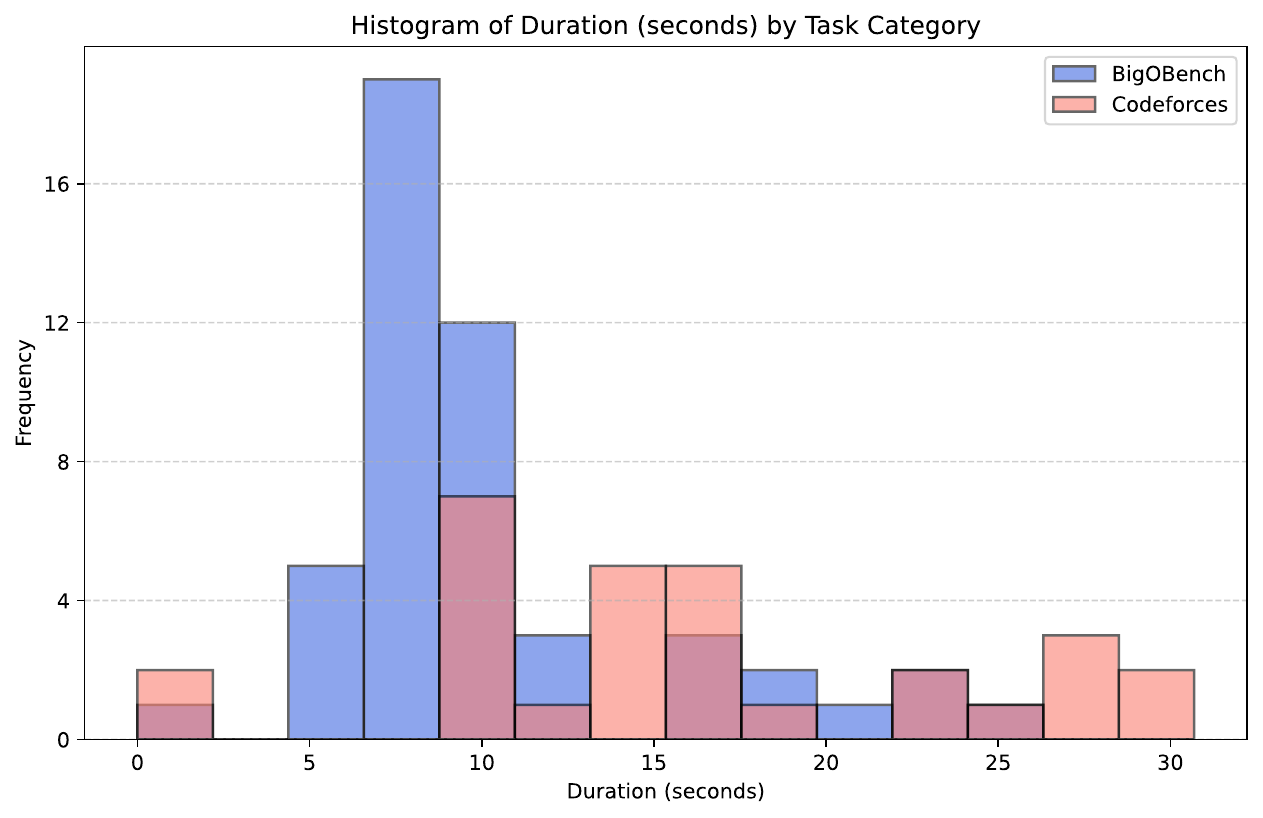}
        \caption{Histogram of the duration (seconds) of each run on the BigO(Bench) and Codeforces datasets.}
        \label{fig:example1}
    \end{subfigure}
    \hfill
    \begin{subfigure}{0.48\textwidth}
        \centering
        \includegraphics[width=\linewidth]{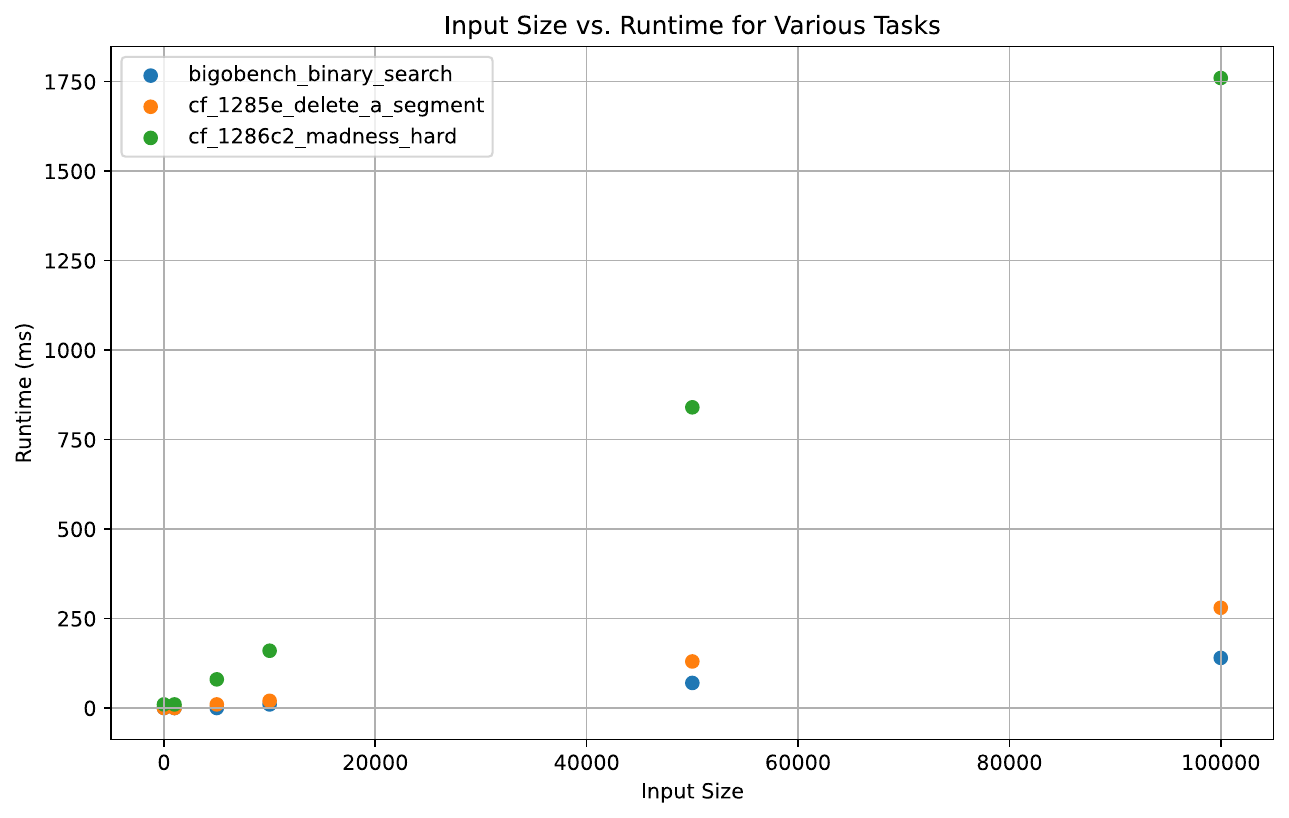}
        \caption{Input size vs. runtime for Big O Bench and Codeforces Div.~2 problems.}
        \label{fig:example2}
    \end{subfigure}

    \caption{Runtime distributions and scaling across benchmarks.}
    \label{fig:combined}
\end{figure}

For each solved task, execution time was recorded over multiple input sizes and fit with an empirical time-complexity model using BigO(Bench). Figure~\ref{fig:example2} shows representative runtime-vs-input-size curves from these measurements. In all cases the measured runtime grew monotonically, matching expected asymptotic behavior (e.g. near-linear or $n\log n$ growth). No abrupt or non-monotonic jumps were observed. 

Notably, runs that required additional replanning tended to incur slightly higher runtimes (reflecting extra profiling and code-synthesis steps), consistent with the small increase in mean durations seen in Section~\ref{sec:mainperform}. The profiling curves confirm that most solutions exhibit polynomial growth and that the system correctly identifies their complexity limits.

\subsection{TLE/MLE Rates}
A total of 21 runs (26.9\% of 78) ended by exceeding contest resource limits. Of these failures, 16 runs (20.5\%) were time-limit exceeded (TLE) and 5 runs (6.4\%) were memory-limit exceeded (MLE). Thus the overall TLE/MLE failure rate was 26.9\%. Figure~\ref{fig:example3} reports these values. This result is consistent with the known sensitivity of CP problems to inefficiency. Most failures were resource-limit violations; in total, we observed 16 TLE and 5 MLE runs. A small number of runs were terminated early by the static pruner (plan rejection).

\subsection{Iteration Count Analysis}
Iterative replanning substantially improved success. Figure~\ref{fig:example4} visualizes these results. Without replanning, 62\% of tasks were solved. Allowing one replanning iteration increased cumulative success to 77\%, and allowing two iterations brought it to 81\%. In other words: 0 replans (initial run only): 62\% tasks solved; 1 replan (two attempts): 77\% cumulative solved; 2 replans (three attempts): 81\% cumulative solved.

Across the \textbf{same} 26 problems, 7 were solved only after one or two replanning steps, indicating that replanning rescued a significant subset of cases. We report cumulative \textbf{Solved@$\le k$} over this identical task list: 0 replans (first attempt only), 1 replan (two attempts), and 2 replans (three attempts). Returns beyond the second replan were marginal. The distribution of iteration counts is plotted in Figure~\ref{fig:example4}.


\subsection{Baselines}
We compare SwiftSolve against two single-pass LLM baselines: direct GPT-4.1 and direct Claude 4 Opus (single-attempt). The GPT-4.1 baseline achieves roughly 60.4\% success on BigO(Bench) and 44.4\% on the Codeforces suite, with average runtimes of about 5.1s and 9.8s respectively. The baseline Claude-Opus is similar: approximately 62. 5\% on BigO and 46.7\% on Codeforces (avg. 5.9s and 5.5s). In contrast, SwiftSolve’s multi-agent approach yields substantially higher success on both datasets: its gains are especially pronounced on the harder Codeforces tasks. These improvements come with a trade-off in speed: SwiftSolve’s structured planning (multiple agent iterations) leads to longer total runtime per problem.

On Codeforces problems, GPT-4.1’s answers ran much slower (9.8s vs 5.5s) than Claude’s, highlighting inefficiency. Because SwiftSolve’s planner uses the Claude 4 Opus model, the Claude baseline provides the most direct comparison. SwiftSolve builds on this by iteratively decomposing problems and pruning plans, which raises overall success (especially on Codeforces) at the cost of higher runtime. Overall, SwiftSolve outperforms both GPT-4.1 and Claude-Opus baselines in solution accuracy while demonstrating the expected runtime/accuracy trade-off inherent in more complex planning.

\section{Discussion}

SwiftSolve's gains stem from coupling algorithmic planning with empirical profiling and complexity analysis as part of the objective rather than a post-hoc property. Iterative replanning \emph{rescues} a meaningful subset of tasks, and dynamic profiling with coarse complexity fitting aligns edits toward the bottleneck rather than blind retries. A lightweight static pruner cheaply filters plans that would deterministically blow time or memory, and the remaining failures are overwhelmingly \textsc{TLE}/\textsc{MLE} errors rather than logic mishaps, indicating computational cost is the principal barrier. Empirically, \textsc{Solved@$\le$3} increases across 0$\to$1$\to$2 replans (61.54$\to$76.92$\to$80.77\%) while mean runtime changes only marginally (11.96$\to$12.58$\to$12.66\,s), indicating improvement without proportional latency cost. Taken together, these signals shift the objective from ``pass once'' to ``pass under resource limits,'' yielding a resource-aware system.

Several directions merit investigation. \textbf{(1) Complexity labeling is threshold sensitive.} The slope–fit procedure discretizes continuous growth and can be unstable at small $n$. Although we quantify accuracy, non-parametric alternatives remain open.  \textbf{(2) Iteration budget and stopping.} We justify a two-replan cap with a budget-sensitivity sweep, but stopping is still heuristic; per-task adaptive budgets and confidence-based halting can be covered in future work. \textbf{(3) External validity.} Results cover two benchmarks, a single language/runtime, and one hardware profile. Future work could include generalization to other languages, toolchains, I/O regimes, and stricter resource policies. \textbf{(4) Attribution and causality.} Beyond the pruner ablation, our analysis is largely observational. We do not localize where efficiency gains arise (e.g., knockout or patch-type causality).

\section{Conclusion}

We introduced an efficiency-aware evaluation that couples standard acceptance metrics (pass@k) with resource-centric measures (eff@k runtime/memory), TLE/MLE incidence, and complexity-fit accuracy. In a snapshot of 78 runs, the overall run success rate was 73. 08\%, and a three-attempt replanning protocol increased the cumulative Solved @ k from 61.54\% to 80.77\% with only marginal latency change. Performance was stronger and faster on BigO(Bench) (77.55\%, 10.30 s) than on the Codeforces-style subset (65.52\%, 15.95 s), showing sensitivity to task structure and difficulty. Profiling-derived signals expose efficiency failure modes that correctness-only metrics can miss. Due to compute limits, the task set is modest; future work will add public baselines, scale datasets and languages, strengthen complexity inference, and evaluate robustness (adversarial inputs, distribution shift). For safety and reproducibility we use sandboxed execution, avoid data leakage, and will pursue gated access, provenance/watermarking, energy reporting, and expanded misuse mitigations.

\section*{Acknowledgments and Disclosure of Funding}
Jonas Rohweder is supported by the Konrad Zuse School of Excellence in Learning and Intelligent
Systems (ELIZA) through the DAAD programme Konrad Zuse Schools of Excellence in Artificial
Intelligence, sponsored by the Federal Ministry of Education and Research.

\section*{Competing interests}
The authors declare no competing interests.

\bibliographystyle{plainnat}
\bibliography{myrefs}

\begin{thebibliography}{19}
\providecommand{\natexlab}[1]{#1}
\providecommand{\url}[1]{\texttt{#1}}
\expandafter\ifx\csname urlstyle\endcsname\relax
  \providecommand{\doi}[1]{doi: #1}\else
  \providecommand{\doi}{doi: \begingroup \urlstyle{rm}\Url}\fi

\bibitem[Chambon et~al.()Chambon, Roziere, Sagot, and
  Synnaeve]{chambon_bigobench_2025}
Pierre Chambon, Baptiste Roziere, Benoit Sagot, and Gabriel Synnaeve.
\newblock {BigO}(bench) -- can {LLMs} generate code with controlled time and
  space complexity?
\newblock URL \url{http://arxiv.org/abs/2503.15242}.

\bibitem[{Codeforces}(2025)]{codeforces-2025}
{Codeforces}.
\newblock {Codeforces}: A competitive programming platform.
\newblock \url{https://codeforces.com}, 2025.
\newblock Accessed: 2025-07-07.

\bibitem[Colvin et~al.(2025)Colvin, Jolibois, Ramezani, Garcia~Badaracco,
  Dorsey, Montague, Matveenko, Trylesinski, Runkle, Hewitt, Hall, and
  Plot]{pydantic_validation}
Samuel Colvin, Eric Jolibois, Hasan Ramezani, Adrian Garcia~Badaracco, Terrence
  Dorsey, David Montague, Serge Matveenko, Marcelo Trylesinski, Sydney Runkle,
  David Hewitt, Alex Hall, and Victorien Plot.
\newblock Pydantic validation, 2025.
\newblock URL \url{https://docs.pydantic.dev/latest/}.

\bibitem[Hong et~al.()Hong, Zhuge, Chen, Zheng, Cheng, Zhang, Wang, Wang, Yau,
  Lin, Zhou, Ran, Xiao, Wu, and Schmidhuber]{hong_metagpt_2024}
Sirui Hong, Mingchen Zhuge, Jiaqi Chen, Xiawu Zheng, Yuheng Cheng, Ceyao Zhang,
  Jinlin Wang, Zili Wang, Steven Ka~Shing Yau, Zijuan Lin, Liyang Zhou, Chenyu
  Ran, Lingfeng Xiao, Chenglin Wu, and Jürgen Schmidhuber.
\newblock {MetaGPT}: Meta programming for a multi-agent collaborative
  framework.
\newblock URL \url{http://arxiv.org/abs/2308.00352}.

\bibitem[Hossain et~al.()Hossain, Tabassum, Arefin, and
  Zaman]{hossain_llm-pros_2025}
Md~Sifat Hossain, Anika Tabassum, Md~Fahim Arefin, and Tarannum~Shaila Zaman.
\newblock {LLM}-{ProS}: Analyzing large language models' performance in
  competitive problem solving.
\newblock URL \url{http://arxiv.org/abs/2502.04355}.

\bibitem[Huang et~al.({\natexlab{a}})Huang, Dai, Weng, Wu, Qing, Cui, Guo, and
  Zhang]{huang_effilearner_2025}
Dong Huang, Jianbo Dai, Han Weng, Puzhen Wu, Yuhao Qing, Heming Cui, Zhijiang
  Guo, and Jie~M. Zhang.
\newblock {EffiLearner}: Enhancing efficiency of generated code via
  self-optimization, {\natexlab{a}}.
\newblock URL \url{http://arxiv.org/abs/2405.15189}.

\bibitem[Huang et~al.({\natexlab{b}})Huang, Zhang, Luck, Bu, Qing, and
  Cui]{huang_agentcoder_2024}
Dong Huang, Jie~M. Zhang, Michael Luck, Qingwen Bu, Yuhao Qing, and Heming Cui.
\newblock {AgentCoder}: Multi-agent-based code generation with iterative
  testing and optimisation, {\natexlab{b}}.
\newblock URL \url{http://arxiv.org/abs/2312.13010}.

\bibitem[Ishibashi and Nishimura()]{ishibashi_self-organized_2024}
Yoichi Ishibashi and Yoshimasa Nishimura.
\newblock Self-organized agents: A {LLM} multi-agent framework toward ultra
  large-scale code generation and optimization.
\newblock URL \url{http://arxiv.org/abs/2404.02183}.

\bibitem[Islam et~al.()Islam, Ali, and Parvez]{islam_codesim_2025}
Md~Ashraful Islam, Mohammed~Eunus Ali, and Md~Rizwan Parvez.
\newblock {CODESIM}: Multi-agent code generation and problem solving through
  simulation-driven planning and debugging.
\newblock URL \url{http://arxiv.org/abs/2502.05664}.

\bibitem[Islam et~al.(2024)Islam, Ali, and
  Parvez]{islam2024mapcodermultiagentcodegeneration}
Md.~Ashraful Islam, Mohammed~Eunus Ali, and Md~Rizwan Parvez.
\newblock Mapcoder: Multi-agent code generation for competitive problem
  solving, 2024.
\newblock URL \url{https://arxiv.org/abs/2405.11403}.

\bibitem[Liu et~al.()Liu, Deng, Kaler, Chen, Leiserson, Ma, and
  Chen]{liu_lessons_2025}
Yuanzhe Liu, Ryan Deng, Tim Kaler, Xuhao Chen, Charles~E. Leiserson, Yao Ma,
  and Jie Chen.
\newblock Lessons learned: A multi-agent framework for code {LLMs} to learn and
  improve.
\newblock URL \url{http://arxiv.org/abs/2505.23946}.

\bibitem[Nichols et~al.()Nichols, Polasam, Menon, Marathe, Gamblin, and
  Bhatele]{nichols_performance-aligned_2024}
Daniel Nichols, Pranav Polasam, Harshitha Menon, Aniruddha Marathe, Todd
  Gamblin, and Abhinav Bhatele.
\newblock Performance-aligned {LLMs} for generating fast code.
\newblock URL \url{http://arxiv.org/abs/2404.18864}.

\bibitem[Pan et~al.()Pan, Zhang, and Liu]{pan_codecor_2025}
Ruwei Pan, Hongyu Zhang, and Chao Liu.
\newblock {CodeCoR}: An {LLM}-based self-reflective multi-agent framework for
  code generation.
\newblock URL \url{http://arxiv.org/abs/2501.07811}.

\bibitem[Qiu et~al.()Qiu, Zeng, Ezick, Lott, and Tong]{qiu_how_2025}
Ruizhong Qiu, Weiliang~Will Zeng, James Ezick, Christopher Lott, and Hanghang
  Tong.
\newblock How efficient is {LLM}-generated code? a rigorous \& high-standard
  benchmark.
\newblock URL \url{http://arxiv.org/abs/2406.06647}.

\bibitem[Qiu et~al.(2025)Qiu, Zeng, Ezick, Lott, and Tong]{qiu2025enamel}
Ruizhong Qiu, Weiliang~Will Zeng, James Ezick, Christopher Lott, and Hanghang
  Tong.
\newblock How efficient is llm-generated code? a rigorous \& high-standard
  benchmark.
\newblock In \emph{International Conference on Learning Representations
  (ICLR)}, 2025.
\newblock URL \url{https://arxiv.org/abs/2406.06647}.
\newblock ENAMEL.

\bibitem[Sepidband et~al.()Sepidband, Taherkhani, Wang, and
  Hemmati]{sepidband_enhancing_2025}
Melika Sepidband, Hamed Taherkhani, Song Wang, and Hadi Hemmati.
\newblock Enhancing {LLM}-based code generation with complexity metrics: A
  feedback-driven approach.
\newblock URL \url{http://arxiv.org/abs/2505.23953}.

\bibitem[Shinn et~al.()Shinn, Cassano, Berman, Gopinath, Narasimhan, and
  Yao]{shinn_reflexion_2023}
Noah Shinn, Federico Cassano, Edward Berman, Ashwin Gopinath, Karthik
  Narasimhan, and Shunyu Yao.
\newblock Reflexion: Language agents with verbal reinforcement learning.
\newblock URL \url{http://arxiv.org/abs/2303.11366}.

\bibitem[Wei et~al.()Wei, Li, Chen, Zheng, Qu, Yu, Chen, and
  Ju]{wei_evaluating_2025}
Minnan Wei, Ziming Li, Xiang Chen, Menglin Zheng, Ziyan Qu, Cheng Yu, Siyu
  Chen, and Xiaolin Ju.
\newblock Evaluating and improving large language models for competitive
  program generation.
\newblock URL \url{http://arxiv.org/abs/2506.22954}.

\bibitem[Zhang et~al.()Zhang, Yang, Bai, Wu, Li, Wang, and
  Li]{zhang_towards_2024}
Yang Zhang, Shixin Yang, Chenjia Bai, Fei Wu, Xiu Li, Zhen Wang, and Xuelong
  Li.
\newblock Towards efficient {LLM} grounding for embodied multi-agent
  collaboration.
\newblock URL \url{http://arxiv.org/abs/2405.14314}.

\end{thebibliography}

\appendix
\section*{Appendix}

\section{Detailed Prompting of SwiftSolve}
\begin{figure}[h]
    \centering
    \includegraphics[width=1.0\textwidth]{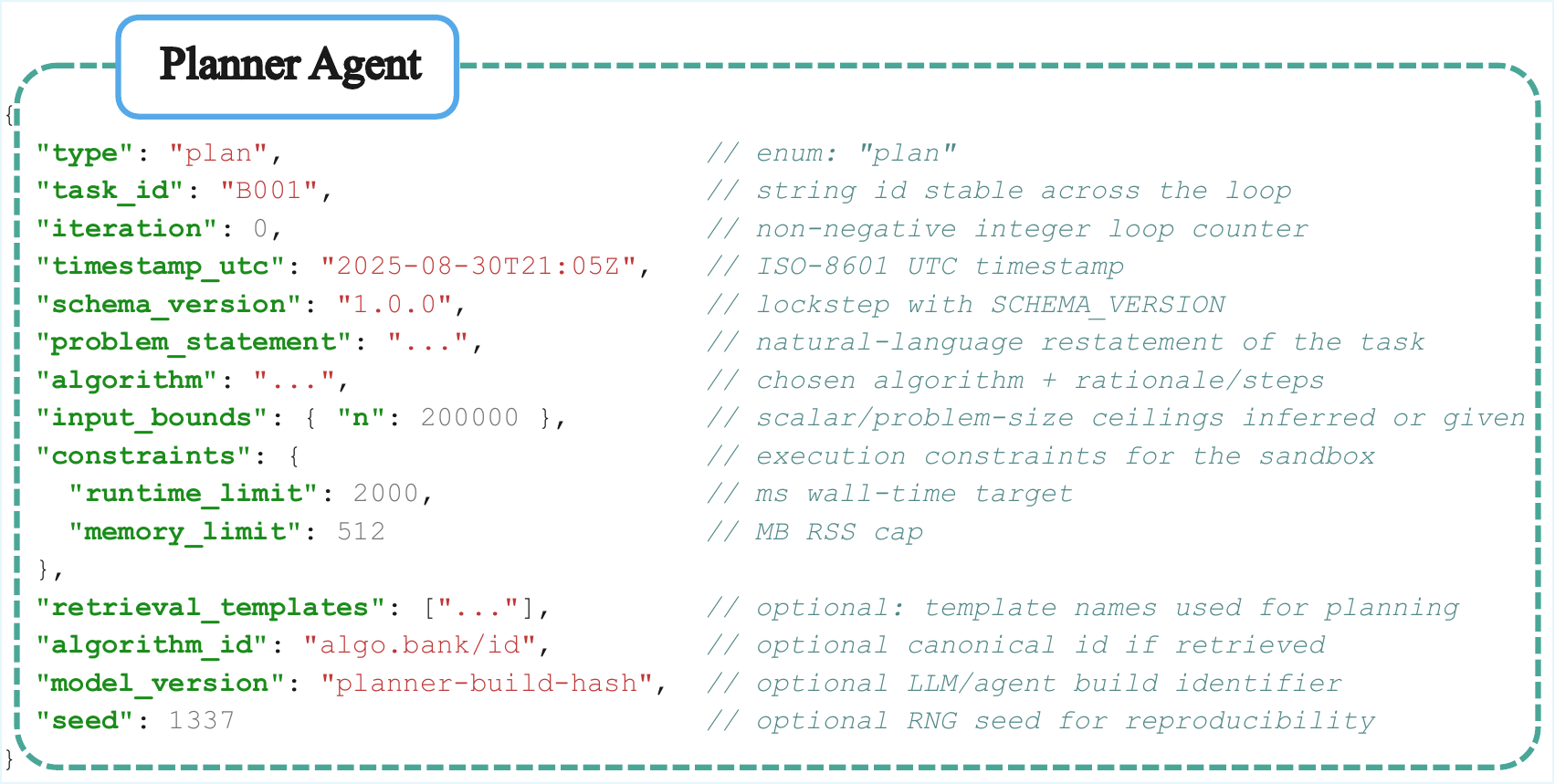}
    \caption{Schema output for the planner agent.}
    \label{fig:planner}
\end{figure}

\begin{figure}[h]
    \centering
    \includegraphics[width=1.0\textwidth]{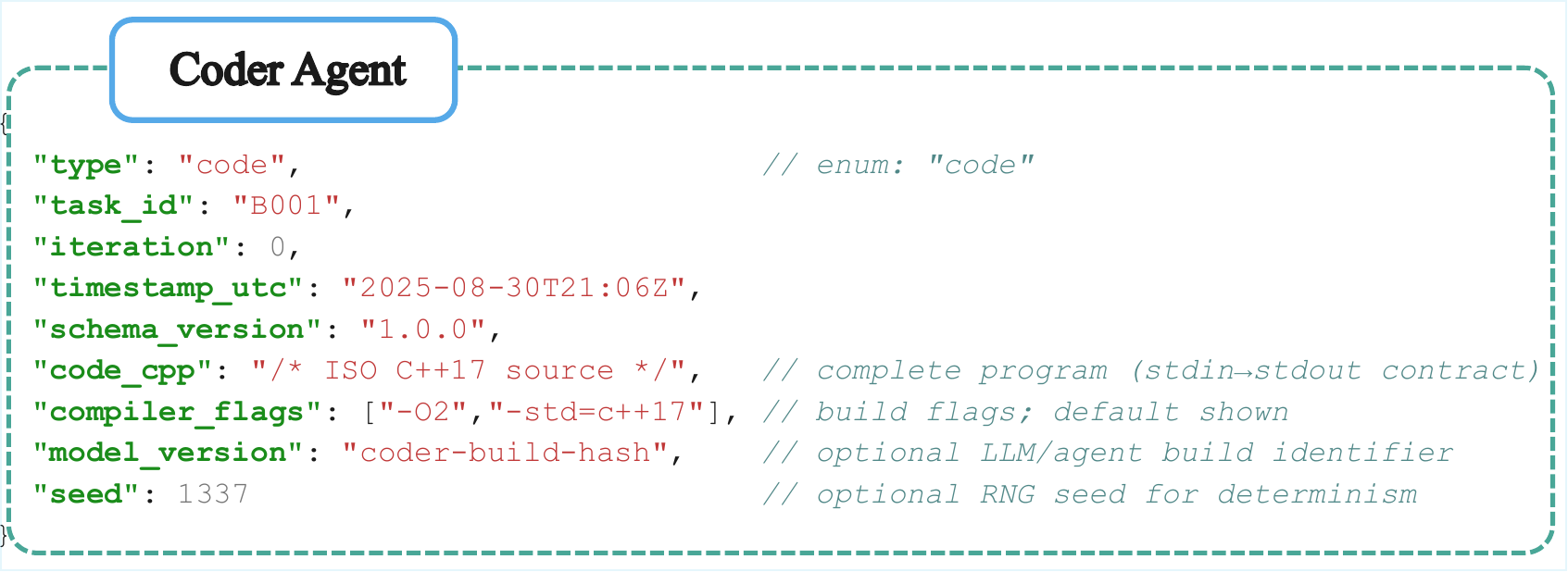}
    \caption{Schema output for the coder agent.}
    \label{fig:coder}
\end{figure}

\begin{figure}[h]
    \centering
    \includegraphics[width=1.0\textwidth]{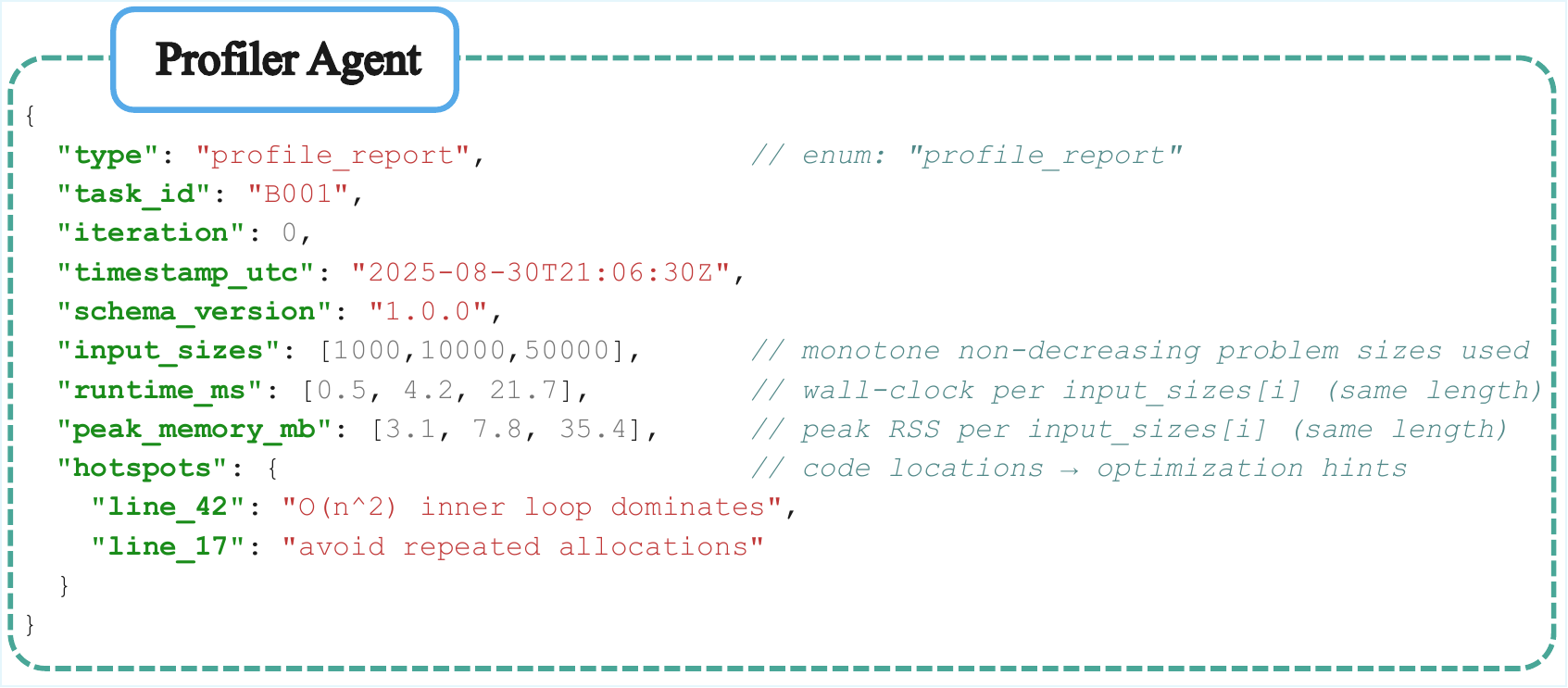}
    \caption{Schema output for the profiler agent.}
    \label{fig:profiler}
\end{figure}

\begin{figure}[h]
    \centering
    \includegraphics[width=1.0\textwidth]{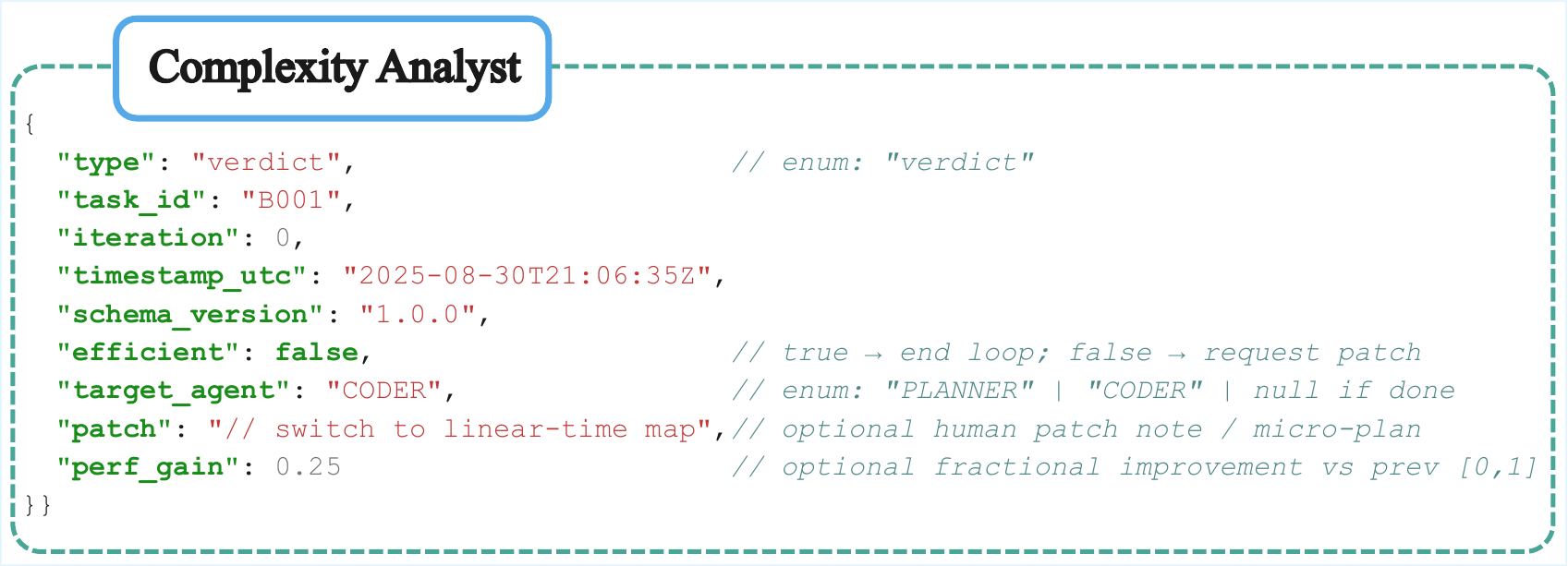}
    \caption{Schema output for the complexity analyst.}
    \label{fig:analyst}
\end{figure}

\begin{figure}[H]
    \centering
    \includegraphics[width=1.0\textwidth]{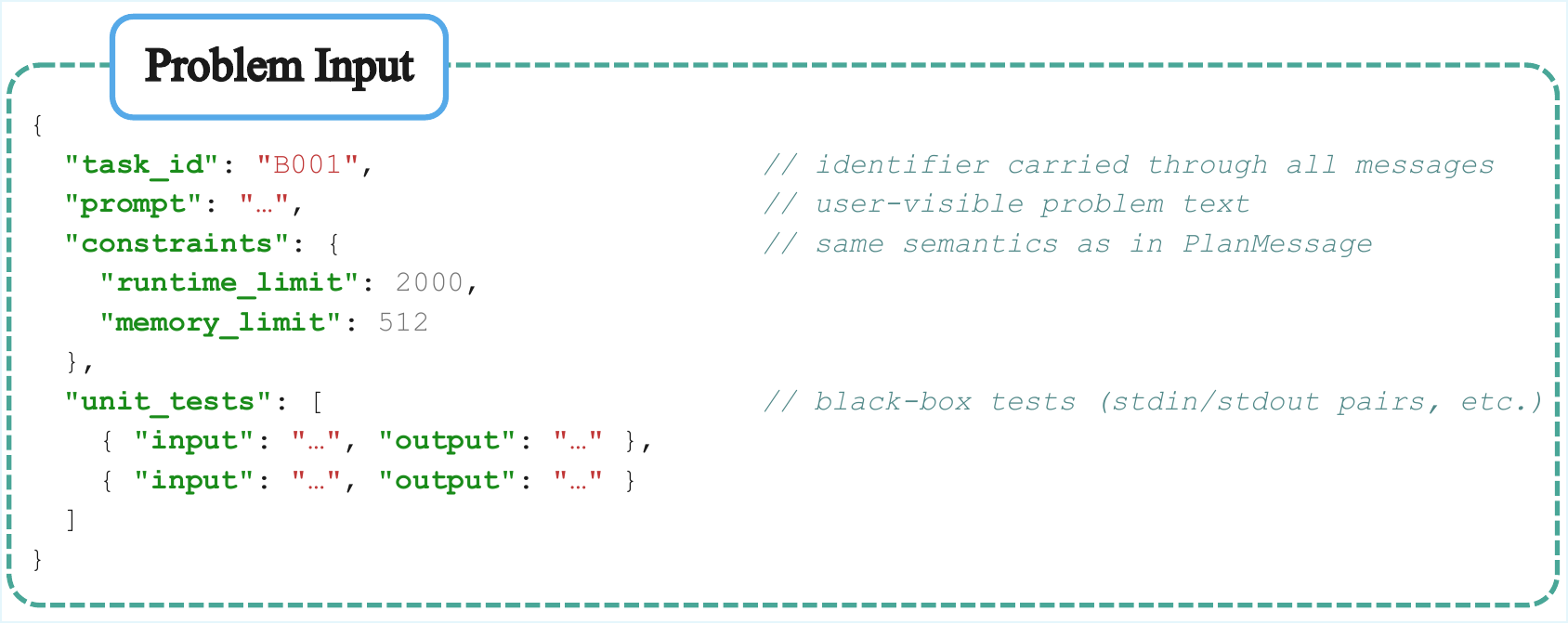}
    \caption{Schema input for competitive programming problems into SwiftSolve.}
    \label{fig:input}
\end{figure}

\begin{figure}[H]
    \centering
    \includegraphics[width=1.0\textwidth]{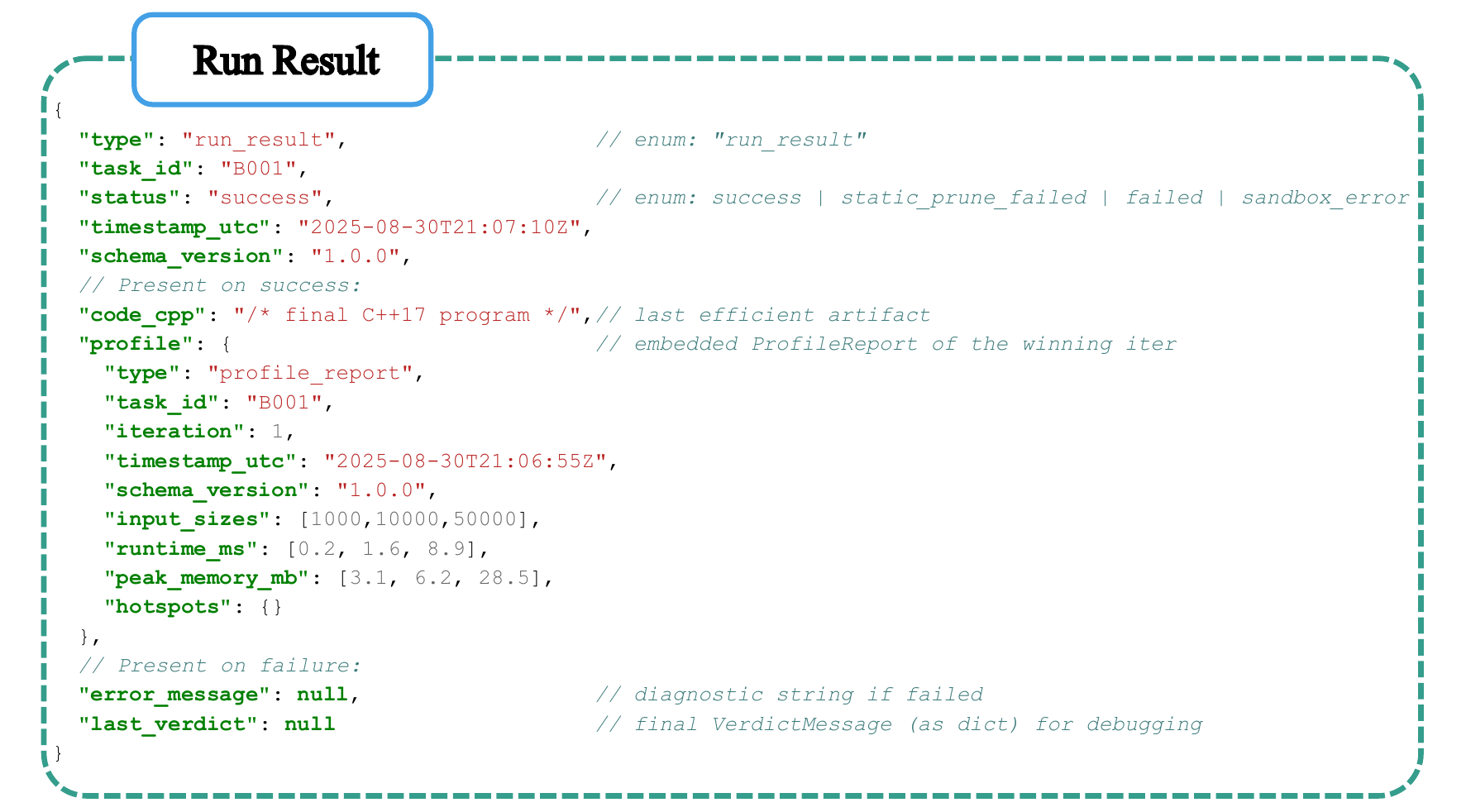}
    \caption{Schema output for competitive programming problems into SwiftSolve.}
    \label{fig:result}
\end{figure}

The detailed prompting schema of the Planner, Coder, Profiler, and Complexity Analyst are shown in Figure~\ref{fig:planner},~\ref{fig:coder},~\ref{fig:profiler}, and~\ref{fig:analyst}, respectively. Additionally, we visualize the competitive programming problem schema input and output for SwiftSolve, noted in Figure~\ref{fig:input} and~\ref{fig:result}, respectively.

\section{Example Dataset Schema}
Example dataset schemas for both BigOBench and Codeforces Div. 2 problems, showing all the tasks we used, can be found at this \href{https://anonymous.4open.science/r/swiftsolve-ED55/datasets/}{link}.

\begin{figure}[H]
    \centering
    \begin{subfigure}{0.48\textwidth}
        \centering
        \includegraphics[width=\linewidth]{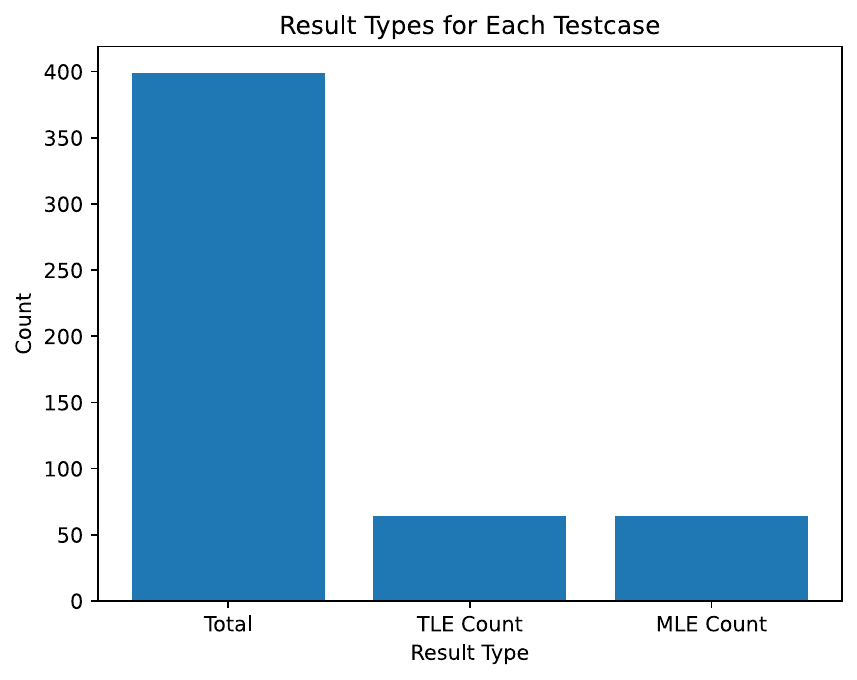}
        \caption{TLE and MLE rates for iterative replanning failures.}
        \label{fig:example3}
    \end{subfigure}
    \hfill
    \begin{subfigure}{0.48\textwidth}
        \centering
        \includegraphics[width=\linewidth]{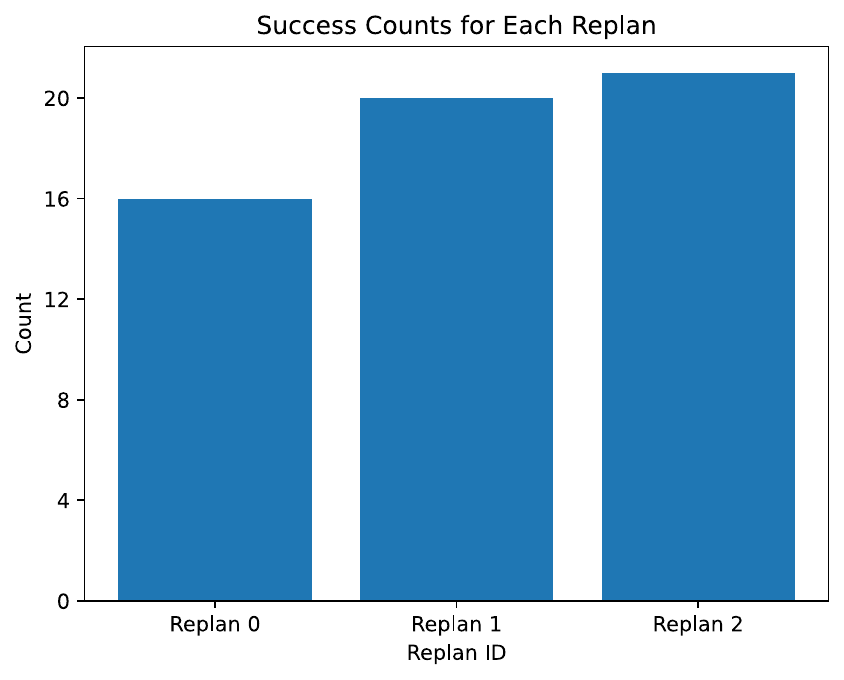}
        \vspace{-3.8mm}
        \caption{Amount of successful iterative replans.}
        \vspace{3.25mm}
        \label{fig:example4}
    \end{subfigure}

    \vspace{-2.0mm}
    \caption{Error incidence and recovery patterns under iterative replanning.}
    \label{fig:combined2}
\end{figure}

\clearpage


\end{document}